\documentclass{article}

\usepackage{colt09e,times}
\usepackage{amsfonts}
\usepackage{amsmath}
\usepackage{mathrsfs}
\usepackage{natbib}
\usepackage{amssymb}
\usepackage{graphicx}

\newif\iflongdoc
\longdoctrue
\makeatletter

\title{Empirical Bernstein Bounds and Sample Variance Penalization}

\author{Andreas Maurer \\
Adalbertstrasse 55 \\ 
D-80799 M\"unchen, Germany \\
{\tt andreasmaurer@compuserve.com} \\
\And Massimiliano Pontil\thanks{This work was partially 
supported by EPSRC Grants GR/T18707/01 and EP/D071542/1.}
\\ Dept. of Computer Science \\
University College London \\
Malet Pl., WC1E,  London, UK \\
{\tt m.pontil@cs.ucl.ac.uk}
}

\begin{document}
\maketitle

\begin{abstract}
We give improved constants for data dependent and variance sensitive
confidence bounds, called empirical Bernstein bounds, and extend these
inequalities to hold uniformly over classes of functions whose growth
function is polynomial in the sample size $n$. The bounds lead us to
consider {\em sample variance penalization}, a novel learning method which
takes into account the empirical variance of the loss function. We
give conditions under which sample variance penalization is
effective. In particular, we present a bound on the excess risk
incurred by the method. Using this, we argue that there are situations
in which the excess risk of our method is of order $1/n$, while the
excess risk of empirical risk minimization is of order
$1/\sqrt{n}$. We show some experimental results, which confirm the
theory. Finally, we discuss the potential application of our results
to sample compression schemes.
\end{abstract}

\section{Introduction}
The method of empirical risk minimization (ERM) is so intuitive, that
some of the less plausible alternatives have received little attention
by the machine learning community. In this work we present
sample variance penalization (SVP), a method which is motivated by
some variance-sensitive, data-dependent confidence bounds, which we
develop in the paper. We describe circumstances under which SVP works
better than ERM and provide some preliminary experimental results
which confirm the theory. 

In order to explain the underlying ideas and highlight the differences
between SVP and ERM, we begin with a discussion of the confidence
bounds most frequently used in learning theory.

\begin{theorem}[Hoeffding's inequality]
\label{Theorem Hoeffdings Inequality}Let $Z,Z_{1},\dots,Z_{n}$ be i.i.d. random
variables with values in $\left[ 0,1\right] $ and let $\delta >0$. Then with
probability at least $1-\delta $ in $\left( Z_{1},\dots,Z_{n}\right) $ we have
\begin{equation*}
\mathbb{E}Z-\frac{1}{n}\sum_{i=1}^{n}Z_{i}\leq \sqrt{\frac{\ln 1/\delta }{2n}
}.
\end{equation*}
\end{theorem}

It is customary to call this result Hoeffding's inequality. It
appears in a stronger, more general form in Hoeffding's 1963
milestone paper \cite{Hoeffding 1963}. Proofs can be found in
\cite{Hoeffding 1963} or \cite{McDiarmid 1998}. We cited
Hoeffding's inequality in form of a confidence-dependent bound on
the deviation, which is more convenient for our discussion than a
deviation-dependent bound on the confidence. Replacing $Z$ by $1-Z$
shows that the confidence interval is symmetric about $\mathbb{E }Z$.

Suppose some underlying observation is modeled by a random variable
$X$, distributed in some space $\mathcal{X}$ according to some
law $\mu $. In learning theory Hoeffding's inequality is
often applied when $Z$ measures the loss incurred by some hypothesis
$h$ when $X$ is observed, that is,
\begin{equation*}
Z=\ell _{h}\left( X\right) .
\end{equation*}
The expectation $\mathbb{E}_{X\sim \mu }\ell _{h}\left( X\right) $ is
called the risk associated with hypothesis $h$ and distribution $\mu
$.  Since the risk depends only on the function $\ell _{h}$ and on
$\mu $ we can write the risk as
\begin{equation*}
P\left( \ell _{h},\mu \right) ,
\end{equation*}
where $P$ is the expectation functional. If an i.i.d. vector $\mathbf{X}=\left(
X_{1},\dots,X_{n}\right) $ has been observed, then Hoeffding's inequality
allows us to estimate the risk, for fixed hypothesis, by the empirical risk 
\begin{equation*}
P_{n}\left( \ell _{h},\mathbf{X}\right) = \frac{1}{n} \sum_{i}\ell
_{h}\left( X_{i}\right)
\end{equation*}
within a confidence interval of length $2\sqrt{\left( \ln 1/\delta \right)
/\left( 2n\right) }$.

Let us call the set $\mathcal{F}$ of functions $\ell _{h}$ for all different
hypotheses $h$ the \textit{hypothesis space} and its members \textit{
hypotheses}, ignoring the distinction between a hypothesis $h$ and the
induced loss function $\ell _{h}$. The bound in Hoeffding's inequality
can easily be adjusted to hold uniformly over any finite hypothesis space $
\mathcal{F}$ to give the following well known result \cite{Anthony 1999}.

\begin{corollary}
\label{Corollary Hoeffding finite function class}Let $X$ be a random
variable with values in a set $\mathcal{X}$ with distribution $\mu $, and
let $\mathcal{F}$ be a finite class of hypotheses $f:\mathcal{X\rightarrow }
\left[ 0,1\right] $ and $\delta >0$. Then with probability at least $
1-\delta $ in $\mathbf{X}=\left( X_{1},\dots,X_{n}\right) \sim \mu ^{n}$
\begin{equation*}
P\left( f,\mu \right) -P_{n}\left( f,
\mathbf{X}\right) \leq \sqrt{\frac{\ln \left( \left\vert \mathcal{F}
\right\vert /\delta \right) }{2n}},~~\forall f\in \mathcal{F}\text{, }
\end{equation*}
where $\left\vert \mathcal{F}\right\vert $ is the cardinality of $\mathcal{F}
$.
\end{corollary}

This result can be further extended to hold uniformly over hypothesis
spaces whose complexity can be controlled with different covering
numbers which then appear in place of the cardinality $\left\vert
\mathcal{F}\right\vert $ above. A large body of literature exists on
the subject of such uniform bounds to justify hypothesis selection by
empirical risk minimization, see \cite{Anthony 1999} and references
therein. Given a sample $\mathbf{X}$ and a hypothesis space
$\mathcal{ F}$, empirical risk minimization selects the hypothesis
\begin{equation*}
ERM\left( \mathbf{X}\right) =\arg \min_{f\in \mathcal{F}}P_{n}\left(
f,
\mathbf{X}\right) \text{.}
\end{equation*}

A drawback of Hoeffding's inequality is that the confidence interval is
independent of the hypothesis in question, and always of order $\sqrt{1/n}$,
leaving us with a uniformly blurred view of the hypothesis class. But for
hypotheses of small variance better estimates are possible, such as the
following, which can be derived from what is usually called Bennett's
inequality (see e.g. Hoeffding's paper \cite{Hoeffding 1963}).

\begin{theorem}[Bennett's inequality]
\label{Theorem Bernsteins inequality}Under the conditions of Theorem \ref
{Theorem Hoeffdings Inequality} we have with probability at least $1-\delta $
that
\begin{equation*}
\mathbb{E}Z-\frac{1}{n}\sum_{i=1}^{n}Z_{i}\leq \sqrt{\frac{2\mathbb{V}Z\ln
1/\delta }{n}}+\frac{\ln 1/\delta }{3n},
\end{equation*}
where $\mathbb{V}Z$ is the variance $\mathbb{V}Z=\mathbb{E}\left( Z-\mathbb{E
}Z\right) ^{2}$.
\end{theorem}

The bound is symmetric about $\mathbb{E}Z$ and for large $n$ the confidence
interval is now close to $2\sqrt{\mathbb{V}Z}$ times the confidence interval
in Hoeffding's inequality. A version of this bound which is uniform over
finite hypothesis spaces, analogous to Corollary \ref{Corollary Hoeffding
finite function class}, is easily obtained, involving now for each
hypothesis $h$\ the variance $\mathbb{V}h\left( X\right) $. If $h_{1}$ and $
h_{2}$ are two hypotheses then $2\sqrt{\mathbb{V}h_{1}\left( X\right) }$ and 
$2\sqrt{\mathbb{V}h_{2}\left( X\right) }$ are always less than or equal to $
1 $ but they can also be much smaller, or one of them can be substantially
smaller than the other one. For hypotheses of zero variance the diameter of
the confidence interval decays as $O\left( 1/n\right) $.

Bennett's inequality therefore provides us with estimates of lower accuracy
for hypotheses of large variance, and higher accuracy for hypotheses of
small variance. Given many hypotheses of equal and nearly minimal empirical
risk it seems intuitively safer to select the one whose true risk can be
most accurately estimated (a point to which we shall return). But
unfortunately the right hand side of Bennett's inequality depends on the
unobservable variance, so our view of the hypothesis class remains uniformly
blurred.

\subsection{Main results and SVP algorithm}
We are now ready to describe the main results of the paper, which
provide the motivation for the SVP algorithm. 

Our first result provides a purely data-dependent bound with similar
properties as Bennett's inequality.
\begin{theorem}
\label{Theorem empirical Bernstein bound degree 1}Under the conditions of
Theorem \ref{Theorem Hoeffdings Inequality} we have with probability at
least $1-\delta $ in the i.i.d. vector $\mathbf{Z}=\left(
Z_{1},\dots,Z_{n}\right) $ that
\begin{equation*}
\mathbb{E}Z-\frac{1}{n}\sum_{i=1}^{n}Z_{i}\leq \sqrt{\frac{2V_{n}\left( 
\mathbf{Z}\right) \ln 2/\delta }{n}}+\frac{7\ln 2/\delta }{3\left(
n-1\right) },
\end{equation*}
where $V_{n}\left( \mathbf{Z}\right) $ is the sample variance
\begin{equation*}
V_{n}\left( \mathbf{Z}\right) =\frac{1}{n\left( n-1\right) }\sum_{1\leq
i<j\leq n}\left( Z_{i}-Z_{j}\right) ^{2}.
\end{equation*}
\end{theorem}
We next extend Theorem \ref{Theorem empirical Bernstein bound
degree 1} over a finite function class.
\begin{corollary}
\label{Corollary empirical Bernstein finite function class}Let $X$ be a
random variable with values in a set $\mathcal{X}$ with distribution $\mu $,
and let $\mathcal{F}$ be a finite class of hypotheses $f:\mathcal{
X\rightarrow }\left[ 0,1\right] $. For $\delta >0,$ $n\geq 2$ we have with
probability at least $1-\delta $ in $\mathbf{X}=\left(
X_{1},\dots,X_{n}\right) \sim \mu ^{n}$ that 
\begin{align*}
& P\left( f,\mu \right) -P_{n}\left( f,\mathbf{X}\right) & \leq &
\sqrt{\frac{2V_{n}\left( f,\mathbf{X}\right) \ln \left( 2\left\vert 
\mathcal{F}\right\vert /\delta \right) }{n}}+
\\
& &~&
+ \frac{7\ln \left( 2\left\vert 
\mathcal{F}\right\vert /\delta \right) }{3\left( n-1\right) },~~\forall f\in \mathcal{F},
\end{align*}
where $V_{n}\left( f,\mathbf{X}\right) =V_{n}\left( f\left( X_{1}\right)
,\dots,f\left( X_{n}\right) \right) $.
\end{corollary}

Theorem \ref{Theorem empirical Bernstein bound degree 1} makes the diameter
of the confidence interval observable. The corollary is obtained from a union bound over $\mathcal{F}$,
analogous to Corollary \ref{Corollary Hoeffding finite function
class}, and provides us with a view of the loss class which is blurred
for hypotheses of large sample variance, and more in focus for
hypotheses of small sample variance.

We note that an analogous result to Theorem \ref{Theorem empirical
Bernstein bound degree 1} is given by Audibert et al. \cite{Audibert
2007}. Our technique of proof is new and the bound we derive has 
a slightly better constant. 
Theorem \ref{Theorem empirical Bernstein bound degree 1} itself resembles 
Bernstein's or Bennett's
inequality, in confidence bound form, but in terms of observable
quantities. For this reason it has been called an
\textit{empirical Bernstein bound} in \cite{Audibert 2008a}. In \cite{Audibert 2007} 
Audibert et al. apply their result to the analysis of algorithms for
the multi-armed bandit problem and in \cite{Audibert 2008a} it is used to
derive stopping rules for sampling procedures. We will prove Theorem
\ref{Theorem empirical Bernstein bound degree 1} in Section
\ref{Section EBB and Variance}, together with some useful confidence
bounds on the standard deviation, which may be valuable in their own right.

Our next result extends the uniform estimate in Corollary
\ref{Corollary empirical Bernstein finite function class} to infinite
loss classes whose complexity can be suitably controlled. Beyond the
simple extension involving covering numbers for $\mathcal{F}$ in the
uniform norm $\left\Vert \cdot\right\Vert _{\infty }$, we can use the
following complexity measure, which is also fairly commonplace in the 
machine learning literature \cite{Anthony 1999},
\cite{Guo}.

For $\epsilon >0$, a function class $\mathcal{F}$ and an integer $n$, the
``growth function" $\mathcal{N}_{\infty }\left( \epsilon ,\mathcal{F}
,n\right) $ is defined as 
\begin{equation*}
\mathcal{N}_{\infty }\left( \epsilon ,\mathcal{F},n\right) =\sup_{\mathbf{x}
\in X^{n}}\mathcal{N}\left( \epsilon ,\mathcal{F}\left( \mathbf{x}\right)
,\left\Vert \cdot\right\Vert _{\infty }\right) ,
\end{equation*}
where $\mathcal{F}\left( \mathbf{x}\right) =\left\{ \left( f\left(
x_{1}\right) ,\dots,f\left( x_{n}\right) \right) :f\in \mathcal{F}\right\}
\subseteq \mathbb{R}^{n}$ and for $A\subseteq 
\mathbb{R}^{n}$ the number $\mathcal{N}\left( \epsilon ,A,\left\Vert \cdot\right\Vert
_{\infty }\right) $ is the smallest cardinality $\left\vert
A_{0}\right\vert $ of a set $A_{0}\subseteq A$ such that $A$ is
contained in the union of $\epsilon $-balls centered at points in
$A_{0}$, in the metric induced by $\left\Vert \cdot\right\Vert _{\infty
}$. 

\begin{theorem}
\label{Theorem covering numbers}Let $X$ be a random variable with values in
a set $\mathcal{X}$ with distribution $\mu $ and let $\mathcal{F}$ be a
class of hypotheses $f:\mathcal{X\rightarrow }\left[ 0,1\right] $. Fix $
\delta \in \left( 0,1\right) ,$ $n\geq 16$ and set
\begin{equation*}
\mathcal{M}\left( n\right) =10\mathcal{N}_{\infty }\left( 1/n,\mathcal{F}
,2n\right) \text{.}
\end{equation*}
Then with probability at least $1-\delta $ in the random vector $\mathbf{X}=\left(
X_{1},\dots,X_{n}\right) \sim \mu ^{n}$ we have 
\begin{align*}
& P\left( f,\mu \right) -P_{n}\left( f,\mathbf{X}\right) & \leq & \sqrt{\frac{18V_{n}\left( f,\mathbf{X}\right) \ln \left( \mathcal{M}
\left( n\right) /\delta \right) }{n}} \\
& &~& +\frac{15\ln \left( \mathcal{M}\left(
n\right) /\delta \right) }{n-1},~~\forall f\in \mathcal{F}. 
\end{align*}
\end{theorem}

The structure of this bound is very similar to Corollary 
\ref{Corollary empirical Bernstein finite function class}, with 
$2\left\vert \mathcal{F}\right\vert$ replaced by
$\mathcal{M}(n)$. In a number of practical cases polynomial growth of
$\mathcal{N}_{\infty }\left( 1/n,\mathcal{F},n\right)$ in $n$ has been
established. For instance, we quote \cite[equation~(28)]{Guo}
which states that for the bounded linear functionals in the
reproducing kernel Hilbert space associated with Gaussian kernels one
has $\ln
\mathcal{N}_{\infty }\left( 1/n,\mathcal{F},2n\right) =O\left( \ln
^{3/2}n\right) $. Composition with fixed Lipschitz functions preserves this
property, so we can see that Theorem \ref{Theorem covering numbers} is
applicable to a large family of function classes which occur in machine
learning. We will prove Theorem \ref{Theorem covering numbers} in Section 
\ref{Section covering numbers}.

Since the minimization of uniform upper bounds is frequent practice in
machine learning, one could consider minimizing the bounds in
Corollary \ref{Corollary empirical Bernstein finite function class} or
Theorem \ref{Theorem covering numbers}. This leads to
\textit{sample variance penalization}, a technique which selects the
hypothesis
\begin{equation*}
SVP_{\lambda }\left( \mathbf{X}\right) =\arg \min_{f\in
\mathcal{F}}P_{n}\left( f,\mathbf{X}\right) +\lambda
\sqrt{\frac{V_{n}\left( f,\mathbf{X }\right) }{n}},
\end{equation*}
where $\lambda \geq 0$ is some regularization parameter. For $\lambda =0$ we
recover empirical risk minimization. The last term on the right hand side
can be regarded as a data-dependent regularizer.

Why, and under which circumstances, should sample variance
penalization work better than empirical risk minimization? If two
hypotheses have the same empirical risk, why should we discard the one
with higher sample variance?  After all, the empirical risk of the
high variance hypothesis may be just as much overestimating the true
risk as underestimating it. In Section \ref{Section SVP vs ERM} we
will argue that the decay of the excess risk of sample variance
penalization can be bounded in terms of the variance of an optimal
hypothesis (see Theorem \ref{Theorem excess risk bound}) 
and if there is an optimal hypothesis with zero variance,
then the excess risk decreases as $1/n$. We also give an example of
such a case where the excess risk of empirical risk minimization
cannot decrease faster than $O\left( 1/\sqrt{n}\right) $. We then
report on the comparison of the two algorithms in a toy experiment.

Finally, in Section \ref{sec:SC} we present some preliminary observations concerning 
the application of empirical Bernstein bounds to sample-compression schemes.


\subsection{Notation}
We summarize the notation used throughout the paper. We define the
following functions on the cube $\left[ 0,1\right] ^{n}$, which will
be used throughout. For every $\mathbf{x}=\left( x_{1},\dots,x_{n}\right)
\in \left[ 0,1\right] ^{n}$ we let 
$$
P_{n}\left( \mathbf{x}\right) =\frac{1}{n}\sum_{i=1}^{n}x_{i}
$$
and
$$
V_{n}\left( \mathbf{x}\right) =\frac{1}{n\left( n-1\right) }
\sum_{i,j=1}^{n}\frac{\left( x_{i}-x_{j}\right) ^{2}}{2}.
$$
If $\mathcal{X}$ is some set, $f:\mathcal{X\rightarrow }\left[ 0,1\right] $
and $\mathbf{x}=\left( x_{1},\dots,x_{n}\right) \in \mathcal{X}^{n}$ we write $
f\left( \mathbf{x}\right) =\left( f\left( x_{1}\right) ,\dots,f\left(
x_{n}\right) \right)$, $P_{n}\left( f,\mathbf{x}
\right) =P_{n}\left( f\left( \mathbf{x}\right) \right) $ and $V_{n}\left( f,
\mathbf{x}\right) =V_{n}\left( f\left( \mathbf{x}\right) \right) $.

Questions of measurability will be ignored throughout, if necessary
this is enforced through finiteness assumptions. If $X$ is a real
valued random variable we use $\mathbb{E}X$ and $\mathbb{V}X$ to
denote its expectation and variance, respectively. If $X$ is a random
variable distributed in some set $\mathcal{X}$ according to a
distribution $\mu $, we write $X\sim \mu $.  Product measures are
denoted by the symbols $\times $ or $\prod $, $\mu ^{n}$ is the
$n$-fold product of $\mu $ and the random variable $\mathbf{X}=\left(
X_{1},\dots,X_{n}\right) \sim \mu ^{n}$ is an i.i.d. sample generated from
$\mu $. If $X\sim \mu $ and $f:\mathcal{X\rightarrow \mathbb{R}}$
then we write $P\left( f,\mu \right) =\mathbb{E}_{X\sim \mu }f\left(
X\right) =\mathbb{E}f\left( X\right) $ and $V\left( f,\mu \right)
=\mathbb{V}_{X\sim \mu }f\left( X\right) =\mathbb{V}f\left( X\right)
$.

\section{Empirical Bernstein bounds and variance estimation\label{Section
EBB and Variance}}

In this section, we prove Theorem \ref{Theorem empirical Bernstein bound
degree 1} and some related useful results, in particular concentration
inequalities for the variance of a bounded random variable, (\ref{Inequality
for i.i.d.}) and (\ref{Upper tail})\ below, which may be of independent
interest. For future use we derive our results for the more general case
where the $X_{i}$ in the sample are independent, but not necessarily
identically distributed.

We need two auxiliary results. One is a concentration inequality for
self-bounding random variables (Theorem 13 in \cite{Maurer 2006}):

\begin{theorem}
\label{Theorem concentration}Let $\mathbf{X}=\left( X_{1},\dots,X_{n}\right) $
be a vector of independent random variables with values in some set $
\mathcal{X}$. For $1\leq k\leq n$ and $y\in \mathcal{X}$, we use $\mathbf{X}%
_{y,k}$ to denote the vector obtained from $\mathbf{X}$ by replacing $X_{k}$
by $y$. Suppose that $a\geq 1$ and that $Z=Z\left( \mathbf{X}\right) $
satisfies the inequalities%
\begin{eqnarray}
Z\left( \mathbf{X}\right) -\inf_{y\in \mathcal{X}}Z\left( \mathbf{X}%
_{y,k}\right) &\leq &1,\forall k  \label{concentration contition I} \\
\sum_{k=1}^{n}\left( Z\left( \mathbf{X}\right) -\inf_{y\in \mathcal{X}%
}Z\left( \mathbf{X}_{y,k}\right) \right) ^{2} &\leq &aZ\left( \mathbf{X}%
\right)  \label{concentration contition II}
\end{eqnarray}%
almost surely. Then, for $t>0$,%
\begin{equation*}
\Pr \left\{ \mathbb{E}Z-Z>t\right\} \leq \exp \left( \frac{-t^{2}}{2a\mathbb{%
E}Z}\right) .
\end{equation*}%
If $Z$ satisfies only the self-boundedness condition (\ref{concentration
contition II}) we still have%
\begin{equation*}
\Pr \left\{ Z-\mathbb{E}Z>t\right\} \leq \exp \left( \frac{-t^{2}}{2a\mathbb{%
E}Z+at}\right) .
\end{equation*}
\end{theorem}

The other result we need is a technical lemma on conditional expectations.

\begin{lemma}
\label{Lemma tedious}Let $X$, $Y$ be i.i.d. random variables with values in an
interval $\left[ a,a+1\right] $. Then%
\begin{equation*}
\mathbb{E}_{X}\left[ \mathbb{E}_{Y}\left( X-Y\right) ^{2}\right] ^{2}\leq
\left( 1/2\right) \mathbb{E}\left( X-Y\right) ^{2}.
\end{equation*}
\end{lemma}

\begin{proof}
The right side of the above inequality is of course the variance $\mathbb{E}\left[ X^{2}-XY%
\right] $. One computes%
\begin{equation*}
\mathbb{E}_{X}\left[ \mathbb{E}_{Y}\left( X-Y\right) ^{2}\right] ^{2}=%
\mathbb{E}\left[ X^{4}+3X^{2}Y^{2}-4X^{3}Y\right] .
\end{equation*}%
We therefore have to show that $\mathbb{E}\left[ g\left( X,Y\right) \right]
\geq 0$ where 
\begin{equation*}
g\left( X,Y\right) =X^{2}-XY-X^{4}-3X^{2}Y^{2}+4X^{3}Y
\end{equation*}%
A rather tedious computation gives 
\begin{align*}
& g\left( X,Y\right) +g\left( Y,X\right) = ~~\\
& =X^{2}-XY-X^{4}-3X^{2}Y^{2}+4X^{3}Y + \\
& \hspace{.7truecm} +Y^{2}-XY-Y^{4}-3X^{2}Y^{2}+4Y^{3}X \\
& =\left( X-Y+1\right) \left( Y-X+1\right) \left( Y-X\right) ^{2}.
\end{align*}%
The latter expression is clearly nonnegative, so%
\begin{equation*}
2\left[ \mathbb{E}g\left( X,Y\right) \right] =\mathbb{E}\left[ g\left(
X,Y\right) +g\left( Y,X\right) \right] \geq 0,
\end{equation*}%
which completes the proof.
\end{proof}

When the random variables $X$ and $Y$ are uniformly distributed on a
finite set, $\left\{ x_{1},\dots,x_{n}\right\}$, Lemma \ref{Lemma tedious} 
gives the following useful corollary.
\begin{corollary}
\label{Corollary tedious}Suppose $\left\{ x_{1},\dots,x_{n}\right\} \subset %
\left[ 0,1\right] $. Then%
\begin{equation*}
\frac{1}{n}\sum_{k}\left( \frac{1}{n}\sum_{j}\left( x_{k}-x_{j}\right)
^{2}\right) ^{2}\leq \frac{1}{2n^{2}}\sum_{k,j}\left( x_{k}-x_{j}\right)
^{2}.
\end{equation*}
\end{corollary}


We first establish confidence bounds for the standard deviation.

\begin{theorem}
\label{Theorem realvalued}Let $n\geq 2$ and $\mathbf{X}=\left(
X_{1},\dots,X_{n}\right) $ be a vector of independent random variables with
values in $\left[ 0,1\right] $. Then for $\delta >0$ we have, writing $%
\mathbb{E}V_{n}$ for $\mathbb{E}_{\mathbf{X}}V_{n}\left( \mathbf{X}\right) $,%
\begin{eqnarray}
\Pr \left\{ \sqrt{\mathbb{E}V_{n}}>\sqrt{V_{n}\left( \mathbf{X}\right) }+%
\sqrt{\frac{2\ln 1/\delta }{n-1}}\right\} &\leq &\delta
\label{Stdev upper bound} \\
\Pr \left\{ \sqrt{V_{n}\left( \mathbf{X}\right) }>\sqrt{\mathbb{E}V_{n}}+%
\sqrt{\frac{2\ln 1/\delta }{n-1}}\right\} &\leq &\delta .
\label{Stdev lower bound}
\end{eqnarray}
\end{theorem}

\begin{proof}
Write $Z\left( \mathbf{X}\right) =nV_{n}\left( \mathbf{X}\right) $. Now fix
some $k$ and choose any $y\in \left[ 0,1\right] $. Then 
\begin{align*}
& Z\left( \mathbf{X}\right) -Z\left( \mathbf{X}_{y,k}\right) = \\
& =\frac{1}{n-1}\sum_{j}\left( \left( X_{k}-X_{j}\right) ^{2}-\left(
y-X_{j}\right) ^{2}\right) \\
& \leq \frac{1}{n-1}\sum_{j}\left( X_{k}-X_{j}\right) ^{2}.
\end{align*}%
It follows that $Z\left( \mathbf{X}\right) -\inf_{y\in \Omega }Z\left( 
\mathbf{X}_{y,k}\right) \leq 1$. We also get%
\begin{align*}
& \sum_{k}\left( Z\left( \mathbf{X}\right) -\inf_{y\in \left[ 0,1\right]
}Z\left( \mathbf{X}_{y,k}\right) \right) ^{2} \leq ~~\\
& \leq \sum_{k}\left( \frac{1}{n-1}\sum_{j}\left( X_{k}-X_{j}\right)
^{2}\right) ^{2} \\
& \leq \frac{n^{3}}{\left( n-1\right) ^{2}}\frac{1}{2n^{2}}\sum_{kj}\left(
X_{k}-X_{j}\right) ^{2} \\
& =\frac{n}{n-1}Z\left( \mathbf{X}\right) ,
\end{align*}%
where we applied Corollary \ref{Corollary tedious} to get the second
inequality. It follows that $Z$ satisfies (\ref{concentration contition I})
and (\ref{concentration contition II}) with $a=n/\left( n-1\right) $. From
Theorem \ref{Theorem concentration} and 
$$\Pr \left\{ \pm \mathbb{E}V_{n}\mp
V_{n}\left( \mathbf{X}\right) >s\right\} =\Pr \left\{ \pm \mathbb{E}Z\mp
Z\left( \mathbf{X}\right) >ns\right\}
$$
we can therefore conclude the
following concentration result for the sample variance: For $s>0$%
\begin{eqnarray}
\Pr \left\{ \mathbb{E}V_{n}-V_{n}\left( \mathbf{X}\right) >s\right\} &\leq
&\exp \left( \frac{-\left( n-1\right) s^{2}}{2\mathbb{E}V_{n}}\right)
\label{Inequality for i.i.d.}~~~ \\
\Pr \left\{ V_{n}\left( \mathbf{X}\right) -\mathbb{E}V_{n}>s\right\} &\leq
&\exp \left( \frac{-\left( n-1\right) s^{2}}{2\mathbb{E}V_{n}+s}\right).~~~
\label{Upper tail}
\end{eqnarray}%
From the lower tail bound (\ref{Inequality for i.i.d.}) we obtain with
probability at least $1-\delta $ that%
\begin{equation*}
\mathbb{E}V_{n}-2\sqrt{\mathbb{E}V_{n}}\sqrt{\frac{\ln 1/\delta }{2\left(
n-1\right) }}\leq V_{n}\left( \mathbf{X}\right) .
\end{equation*}%
Completing the square on the left hand side, taking the square-root, adding $%
\sqrt{\ln \left( 1/\delta \right) /\left( 2\left( n-1\right) \right) }$ and
using $\sqrt{a+b}\leq \sqrt{a}+\sqrt{b}$ gives (\ref{Stdev upper bound}).
Solving the right side of (\ref{Upper tail}) for $s$ and using the same
square-root inequality we find that with probability at least $1-\delta $ we
have%
\begin{eqnarray*}
V_{n}\left( \mathbf{X}\right) &\leq &\mathbb{E}V_{n}+2\sqrt{\frac{\mathbb{E}%
V_{n}\ln 1/\delta }{2\left( n-1\right) }}+\frac{\ln 1/\delta }{\left(
n-1\right) } \\
&=&\left( \sqrt{\mathbb{E}V_{n}}+\sqrt{\frac{\ln 1/\delta }{2\left(
n-1\right) }}\right) ^{2}+\frac{\ln 1/\delta }{2\left( n-1\right) }.
\end{eqnarray*}%
Taking the square-root and using the root-inequality again gives (\ref{Stdev
lower bound}).
\end{proof}

We can now prove the empirical Bernstein bound, which reduces to Theorem \ref%
{Theorem empirical Bernstein bound degree 1} for identically distributed
variables.

\begin{theorem}
\label{Theorem Bernstein Variance}Let $\mathbf{X}=\left(
X_{1},\dots,X_{n}\right) $ be a vector of independent random variables with
values in $\left[ 0,1\right] $. Let $\delta >0$. Then with probability at
least $1-\delta $ in $\mathbf{X}$\ we have%
\begin{equation*}
\mathbb{E}\left[ P_{n}\left( \mathbf{X}\right) \right] \leq P_{n}\left( 
\mathbf{X}\right) +\sqrt{\frac{2V_{n}\left( \mathbf{X}\right) \ln 2/\delta }{%
n}}+\frac{7\ln 2/\delta }{3\left( n-1\right) }.
\end{equation*}
\end{theorem}

\begin{proof}
Write $W=\left( 1/n\right) \sum_{i}\mathbb{V}X_{i}$ and observe that 
\begin{eqnarray}
W &\leq &\frac{1}{n}\sum_{i}\mathbb{E}\left( X_{i}-\mathbb{E}X_{i}\right)
^{2}  \label{Variance bound} \\
&&+\frac{1}{2n\left( n-1\right) }\sum_{i\neq j}\left( \mathbb{E}X_{i}-%
\mathbb{E}X_{j}\right) ^{2} \\
&=&\frac{1}{2n\left( n-1\right) }\sum_{i,j}\mathbb{E}\left(
X_{i}-X_{j}\right) ^{2}  \notag \\
&=&\mathbb{E}V_{n}.
\end{eqnarray}%
Recall that Bennett's inequality, which holds also if the $X_{i}$ are not
identically distributed (see \cite{McDiarmid 1998}), implies with
probability at least $1-\delta $ 
\begin{eqnarray*}
\mathbb{E}P_{n}\left( \mathbf{X}\right)  &\leq &P_{n}\left( \mathbf{X}%
\right) +\sqrt{\frac{2W\ln 1/\delta }{n}}+\frac{\ln 1/\delta }{3n} \\
&\leq &P_{n}\left( \mathbf{X}\right) +\sqrt{\frac{2\mathbb{E}V_{n}\ln
1/\delta }{n}}+\frac{\ln 1/\delta }{3n},
\end{eqnarray*}%
so that the conclusion follows from combining this inequality with (\ref%
{Stdev upper bound}) in a union bound and some simple estimates.
\end{proof}

\section{Empirical Bernstein bounds for function classes of polynomial
growth \label{Section covering numbers}}

We now prove Theorem \ref{Theorem covering numbers}. We will use the
classical double-sample method (\cite{Vapnik 1995}, \cite{Anthony 1999}),
but we have to pervert it somewhat to adapt it to the nonlinearity of the
empirical standard-deviation functional. Define functions $\Phi $, $\Psi :%
\left[ 0,1\right] ^{n}\times 
\mathbb{R}
_{+}\rightarrow 
\mathbb{R}
$ by 
\begin{eqnarray*}
\Phi \left( \mathbf{x},t\right) &=&P_{n}\left( \mathbf{x}\right) +\sqrt{%
\frac{2V_{n}\left( \mathbf{x}\right) t}{n}}+\frac{7t}{3\left( n-1\right) }, \\
\Psi \left( \mathbf{x},t\right) &=&P_{n}\left( \mathbf{x}\right) +\sqrt{%
\frac{18V_{n}\left( \mathbf{x}\right) t}{n}}+\frac{11t}{n-1}.
\end{eqnarray*}

We first record some simple Lipschitz properties of these functions.

\begin{lemma}
\label{Lemma Lipschitz}For $t>0$, $\mathbf{x},\mathbf{x}^{\prime }\in \left[
0,1\right] ^{n}$ we have
\begin{eqnarray}
\nonumber
(i)~~ \Phi \left( \mathbf{x},t\right) -\Phi \left( \mathbf{x}^{\prime
},t\right) &\leq& \left( 1+2\sqrt{t/n}\right) \left\Vert \mathbf{x}-\mathbf{x}
^{\prime }\right\Vert _{\infty },\\
(ii)~~ \Psi \left( \mathbf{x},t\right) -\Psi \left( \mathbf{x}^{\prime
},t\right) &\leq& \left( 1+6\sqrt{t/n}\right) \left\Vert \mathbf{x}-\mathbf{x}%
^{\prime }\right\Vert _{\infty }.
\nonumber
\end{eqnarray}
\end{lemma}

\begin{proof}
One verifies that 
$$
\sqrt{V_{n}\left( \mathbf{x}\right) }-\sqrt{V_{n}\left( 
\mathbf{x}^{\prime }\right) }\leq \sqrt{2}\left\Vert \mathbf{x}-\mathbf{x}%
^{\prime }\right\Vert _{\infty },
$$ which implies (i) and (ii).
\end{proof}

Given two vectors $\mathbf{x},\mathbf{x}^{\prime }\in \mathcal{X}^{n}$ and $%
\mathbf{\sigma }\in \left\{ -1,1\right\} ^{n}$ define $\left( \mathbf{\sigma 
},\mathbf{x},\mathbf{x}^{\prime }\right) \in \mathcal{X}^{n}$ by $\left( 
\mathbf{\sigma },\mathbf{x},\mathbf{x}^{\prime }\right) _{i}=x_{i}$ if $%
\sigma _{i}=1$ and $\left( \mathbf{\sigma },\mathbf{x},\mathbf{x}^{\prime
}\right) _{i}=x_{i}^{\prime }$ if $\sigma _{i}=-1$. In the following the $%
\sigma _{i}$ will be independent random variables, uniformly distributed on $%
\left\{ -1,1\right\} $.

\begin{lemma}
\label{Lemma symmetrization} Let $\mathbf{X}=\left( X_{1},\dots,X_{n}\right) $
and $\mathbf{X}^{\prime }=\left( X_{1}^{\prime },\dots,X_{n}^{\prime }\right) $
be random vectors with values in $\mathcal{X}$ such that all the $X_{i}$ and 
$X_{i}^{\prime }$ are independent and identically distributed. Suppose that $%
F:\mathcal{X}^{2n}\rightarrow \left[ 0,1\right] $. Then 
\begin{equation*}
\mathbb{E}F\left( \mathbf{X},\mathbf{X}^{\prime }\right) \leq \sup_{\left( 
\mathbf{x},\mathbf{x}^{\prime }\right) \in \mathcal{X}^{2n}}\mathbb{E}%
_{\sigma }F\left( \left( \mathbf{\sigma },\mathbf{x},\mathbf{x}^{\prime
}\right) ,\left( -\mathbf{\sigma },\mathbf{x},\mathbf{x}^{\prime }\right)
\right) .
\end{equation*}
\end{lemma}

\begin{proof}
For any configuration $\mathbf{\sigma }$ and $\left( \mathbf{X},\mathbf{X}%
^{\prime }\right) $, the configuration $\left( \left( \mathbf{\sigma },%
\mathbf{X},\mathbf{X}^{\prime }\right) ,\left( -\mathbf{\sigma },\mathbf{X},%
\mathbf{X}^{\prime }\right) \right) $ is obtained from $\left( \mathbf{X},%
\mathbf{X}^{\prime }\right) $ by exchanging $X_{i}$ and $X_{i}^{\prime }$
whenever $\sigma _{i}=-1$. Since $X_{i}$ and $X_{i}^{\prime }$ are
identically distributed this does not affect the expectation. Thus%
\begin{eqnarray*}
\mathbb{E}F\left( \mathbf{X},\mathbf{X}^{\prime }\right) &=&\mathbb{E}%
_{\sigma }\mathbb{E}F\left( \left( \mathbf{\sigma },\mathbf{X},\mathbf{X}%
^{\prime }\right) ,\left( -\mathbf{\sigma },\mathbf{X},\mathbf{X}^{\prime
}\right) \right) \\
&\leq &\sup_{\left( \mathbf{x},\mathbf{x}^{\prime }\right) \in \mathcal{X}%
^{2n}}\mathbb{E}_{\sigma }F\left( \left( \mathbf{\sigma },\mathbf{x},\mathbf{%
x}^{\prime }\right) ,\left( -\mathbf{\sigma },\mathbf{x},\mathbf{x}^{\prime
}\right) \right) .
\end{eqnarray*}
\end{proof}

The next lemma is where we use the concentration results in Section \ref%
{Section EBB and Variance}.

\begin{lemma}
\label{Lemma probability bound}Let $f:\mathcal{X}\rightarrow \left[ 0,1%
\right] $ and $\left( \mathbf{x},\mathbf{x}^{\prime }\right) \in \mathcal{X}%
^{2n}$ be fixed. Then%
\begin{equation*}
\Pr_{\sigma }\left\{ \Phi \left( f\left( \mathbf{\sigma },\mathbf{x},\mathbf{%
x}^{\prime }\right) ,t\right) >\Psi \left( f\left( -\mathbf{\sigma },\mathbf{%
x},\mathbf{x}^{\prime }\right) ,t\right) \right\} \leq 5e^{-t}.
\end{equation*}
\end{lemma}

\begin{proof}
Define the random vector $\mathbf{Y}=\left( Y_{1},\dots,Y_{n}\right) $, where
the $Y_{i}$ are independent random variables, each $Y_{i}$ being uniformly
distributed on $\left\{ f\left( x_{i}\right) ,f\left( x_{i}^{\prime }\right)
\right\} $. The $Y_{i}$ are of course not identically distributed. Within
this proof we use the shorthand notation $\mathbb{E}P_{n}=\mathbb{E}_{%
\mathbf{Y}}P_{n}\left( \mathbf{Y}\right) $ and $\mathbb{E}V_{n}=\mathbb{E}_{%
\mathbf{Y}}V_{n}\left( \mathbf{Y}\right) $, and let 
\begin{equation*}
A=\mathbb{E}P_{n}+\sqrt{\frac{8\mathbb{E}V_{n}~t}{n}}+\frac{14t}{3\left(
n-1\right) }.
\end{equation*}%
Evidently
\begin{eqnarray*}
&&\Pr_{\sigma }\left\{ \Phi \left( f\left( \mathbf{\sigma },\mathbf{x},
\mathbf{x}^{\prime }\right) ,t\right) >\Psi \left( f\left( -\mathbf{\sigma },
\mathbf{x},\mathbf{x}^{\prime }\right) ,t\right) \right\} \leq \\
&& \leq \Pr_{\sigma }\left\{ \Phi \left( f\left( \mathbf{\sigma },\mathbf{x},
\mathbf{x}^{\prime }\right) ,t\right) >A\right\}+ 
\\ && \hspace{1.9truecm}
+\Pr_{\sigma }\left\{
A>\Psi \left( f\left( -\mathbf{\sigma },\mathbf{x},\mathbf{x}^{\prime
}\right) ,t\right) \right\} = \\
&& = \Pr_{\mathbf{Y}}\left\{ \Phi \left( \mathbf{Y},t\right) >A\right\} +\Pr_{
\mathbf{Y}}\left\{ A>\Psi \left( \mathbf{Y},t\right) \right\} .
\end{eqnarray*}
To prove our result we will bound these two probabilities in turn.

Now
\begin{eqnarray*}
&&\Pr_{\mathbf{Y}}\left\{ \Phi \left( \mathbf{Y},t\right) >A\right\} \leq \\
&\leq &\Pr \left\{ P_{n}\left( \mathbf{Y}\right) >\mathbb{E}P_{n}+\sqrt{%
\frac{2\mathbb{E}V_{n}t}{n}}+\frac{t}{3\left( n-1\right) }\right\} + \\
&&+\Pr \left\{ \sqrt{\frac{2V_{n}\left( \mathbf{Y}\right) t}{n}}>\sqrt{\frac{%
2\mathbb{E}V_{n}~t}{n}}+\frac{2t}{n-1}\right\} .
\end{eqnarray*}%
Since $\sum_{i}\mathbb{V}\left( f\left( Y_{i}\right) \right) \leq n\mathbb{E}%
V_{n}$ by equation (\ref{Variance bound}), the first of these probabilities
is at most $e^{-t}$ by Bennett's inequality, which also holds for variables
which are not identically distributed. That the second of these
probabilities is bounded by $e^{-t}$ follows directly from Theorem \ref%
{Theorem realvalued} (\ref{Stdev lower bound}). We conclude that $\Pr_{%
\mathbf{Y}}\left\{ \Phi \left( \mathbf{Y},t\right) >A\right\} \leq 2e^{-t}$.

Since $\sqrt{2}+\sqrt{8}=\sqrt{18}$ we have%
\begin{eqnarray*}
&&\Pr_{\mathbf{Y}}\left\{ A>\Psi \left( \mathbf{Y},t\right) \right\} \leq \\
&\leq &\Pr \left\{ \mathbb{E}P_{n}>P_{n}\left( \mathbf{Y}\right) +\sqrt{%
\frac{2V_{n}\left( \mathbf{Y}\right) t}{n}}+\frac{7t}{3\left( n-1\right) }%
\right\} + \\
&&+\Pr \left\{ \sqrt{\frac{8\mathbb{E}V_{n}~t}{n}}>\sqrt{\frac{8V_{n}\left( 
\mathbf{Y}\right) t}{n}}+\frac{4t}{n-1}\right\} .
\end{eqnarray*}%
The first probability in the sum is at most $2e^{-t}$ by Theorem \ref%
{Theorem Bernstein Variance}, and the second is at most $e^{-t}$ by Theorem %
\ref{Theorem realvalued} (\ref{Stdev upper bound}). Hence $\Pr_{\mathbf{Y}%
}\left\{ A>\Psi \left( f\left( \mathbf{Y}\right) ,t\right) \right\} \leq
3e^{-t}$, so it follows that 
\begin{equation*}
\Pr_{\sigma }\left\{ \Phi \left( f\left( \mathbf{\sigma },\mathbf{x},\mathbf{%
x}^{\prime }\right) ,t\right) >\Psi \left( f\left( -\mathbf{\sigma },\mathbf{%
x},\mathbf{x}^{\prime }\right) ,t\right) \right\} \leq 5e^{-t}\text{.}
\end{equation*}
\end{proof}

\noindent {\bf Proof of Theorem \protect\ref{Theorem covering numbers}}. It follows from Theorem \ref{Theorem Bernstein Variance} that for $t>\ln 4$
we have for any $f\in \mathcal{F}$ that 
\begin{equation*}
\Pr \left\{ \Phi \left( f\left( \mathbf{X}\right) ,t\right) >P\left( f,\mu
\right) \right\} \geq 1/2\text{. }
\end{equation*}%
In other words, the functional 
\begin{equation*}
f\mapsto \Lambda \left( f\right) =\mathbb{E}_{\mathbf{X}^{\prime }}\mathbf{1}%
\left\{ \Phi \left( f\left( \mathbf{X}^{\prime }\right) ,t\right) >P\left(
f,\mu \right) \right\} 
\end{equation*}%
satisfies $1\leq 2\Lambda \left( f\right) $ for all $f$. Consequently, for
any $s>0$ we have, using $\mathbb{I}A$ to denote the indicator function of $A
$, that
\begin{align*}
& \Pr_{\mathbf{X}}\left\{ \exists f\in \mathcal{F}:P\left( f,\mu \right)
>\Psi \left( f\left( \mathbf{X}\right) ,t\right) +s\right\}  \\ \\
& =\mathbb{E}_{\mathbf{X}}\sup_{f\in \mathcal{F}}\mathbb{I}\left\{ P\left(
f,\mu \right) >\Psi \left( f\left( \mathbf{X}\right) ,t\right) +s\right\}  \\ \\
& \leq \mathbb{E}_{\mathbf{X}}\sup_{f\in \mathcal{F}}\mathbb{I}\left\{
P\left( f,\mu \right) >\Psi \left( f\left( \mathbf{X}\right) ,t\right)
+s\right\} 2\Lambda \left( f\right)  \\ \\
& =2\mathbb{E}_{\mathbf{X}}\sup_{f\in \mathcal{F}}\mathbb{E}_{\mathbf{X}%
^{\prime }}\mathbb{I}\left\{ 
\begin{array}{c}
P\left( f,\mu \right) >\Psi \left( f\left( \mathbf{X}\right) ,t\right) +s%
\text{ }  \\
\text{and }\Phi \left( f\left( \mathbf{X}^{\prime }\right) ,t\right)
>P\left( f,\mu \right) 
\end{array}%
\right\}  \\ \\
& \leq 2\mathbb{E}_{\mathbf{XX}^{\prime }}\sup_{f\in \mathcal{F}}\mathbb{I}%
\left\{ 
\begin{array}{c}
P\left( f,\mu \right) >\Psi \left( f\left( \mathbf{X}\right) ,t\right) +s%
\text{ }   \\
\text{and }\Phi \left( f\left( \mathbf{X}^{\prime }\right) ,t\right)
>P\left( f,\mu \right) 
\end{array}%
\right\}  \\ \\
& \leq 2\mathbb{E}_{\mathbf{XX}^{\prime }}\sup_{f\in \mathcal{F}}\mathbb{I}
\left\{ \Phi \left( f\left( \mathbf{X}^{\prime }\right) ,t\right) >\Psi
\left( f\left( \mathbf{X}\right) ,t\right) +s\right\}  \\ \\
& \leq 2\sup_{\left( \mathbf{x},\mathbf{x}^{\prime }\right) \in \mathcal{X}
^{2n}}\Pr_{\sigma }\big\{ \exists f\in \mathcal{F}:\Phi \left( f\left( 
\mathbf{\sigma },\mathbf{x},\mathbf{x}^{\prime }\right) ,t\right)\\ 
& \hspace{3truecm} > \Psi \left( f\left( -\mathbf{\sigma },\mathbf{x},\mathbf{x}^{\prime }\right)
,t\right) +s\big\},
\end{align*}%

\noindent where we used Lemma \ref{Lemma symmetrization} in the last step.

Now we fix $\left( \mathbf{x},\mathbf{x}^{\prime }\right) \in \mathcal{X}%
^{2n}$ and let $\epsilon >0$ be arbitrary. We can choose a finite subset $
\mathcal{F}_{0}$ of $\mathcal{F}$ such that $\left\vert \mathcal{F}%
_{0}\right\vert \leq \mathcal{N}\left( \epsilon ,\mathcal{F},2n\right) $ and
that $\forall f\in \mathcal{F}$ there exists $\hat{f}\in \mathcal{F}_{0}$ such
that $\left\vert f\left( x_{i}\right) -\hat{f}\left( x_{i}\right)
\right\vert <\epsilon $ and $\left\vert f\left( x_{i}^{\prime }\right) -\hat{%
f}\left( x_{i}^{\prime }\right) \right\vert <\epsilon $, for all $i\in
\left\{ 1,\dots,n\right\} $. Suppose there exists $f\in \mathcal{F}$ such that 
\begin{equation*}
\Phi \left( f\left( \mathbf{\sigma },\mathbf{x},\mathbf{x}^{\prime }\right)
,t\right) >\Psi \left( f\left( -\mathbf{\sigma },\mathbf{x},\mathbf{x}%
^{\prime }\right) ,t\right) +\left( 2+8\sqrt{\frac{t}{n}}\right) \epsilon 
\text{. }
\end{equation*}%
It follows from the Lemma \ref{Lemma Lipschitz} (i) and (ii) that
there must exist $\hat{f}\in \mathcal{F}_{0}$ such that 
$$\Phi \left( \hat{f}%
\left( \mathbf{\sigma },\mathbf{x},\mathbf{x}^{\prime }\right) ,t\right)
>\Psi \left( \hat{f}\left( -\mathbf{\sigma },\mathbf{x},\mathbf{x}^{\prime
}\right) ,t\right). 
$$
We conclude from the above that%
\begin{align*}
& \Pr_{\sigma }\left\{ 
\begin{array}{c}
\exists f\in \mathcal{F}:\Phi \left( f\left( \mathbf{\sigma },\mathbf{x},%
\mathbf{x}^{\prime }\right) ,t\right) > \\ 
>\Psi \left( f\left( -\mathbf{\sigma },\mathbf{x},\mathbf{x}^{\prime
}\right) ,t\right) +\left( 2+8\sqrt{\frac{t}{n}}\right) \epsilon 
\end{array}%
\right\}  \\ \\
& \leq \Pr_{\sigma }\big\{ \exists f\in \mathcal{F}_{0}:\Phi \left( f\left( 
\mathbf{\sigma },\mathbf{x},\mathbf{x}^{\prime }\right) ,t\right) >\Psi
\left( f\left( -\mathbf{\sigma },\mathbf{x},\mathbf{x}^{\prime }\right)
,t\right) \big\}  \\ 
& \leq \sum_{f\in \mathcal{F}_{0}}\Pr_{\sigma }\big\{ \Phi \left( f\left( 
\mathbf{\sigma },\mathbf{x},\mathbf{x}^{\prime }\right) ,t\right) >\Psi
\left( f\left( -\mathbf{\sigma },\mathbf{x},\mathbf{x}^{\prime }\right)
,t\right) \big\}  \\ 
& \leq 5\mathcal{N}\left( \epsilon ,\mathcal{F},2n\right) e^{-t}\text{,}
\end{align*}%

\noindent where we used Lemma \ref{Lemma probability bound} in the last step. We
arrive at the statement that%
\begin{align*}
& \Pr_{\mathbf{X}}\left\{ \exists f\in \mathcal{F}:P\left( f,\mu \right)
\geq \Psi \left( f\left( \mathbf{X}\right) ,t\right) +\left( 2+8\sqrt{\frac{t%
}{n}}\right) \epsilon \right\}  \\
& \leq 10\mathcal{N}\left( \epsilon ,\mathcal{F},2n\right) e^{-t}.
\end{align*}
Equating this probability to $\delta $, solving for $t$, substituting $
\epsilon =1/n$ and using $8\sqrt{t/n}\leq 2t$, for $n\geq 16$ and $t\geq 1$,
give the result. \qed

\vspace{.3truecm}

We remark that a simplified version of the above argument gives uniform
bounds for the standard deviation $\sqrt{V\left( f,\mu \right) }$, using
Theorem \ref{Theorem realvalued} (\ref{Stdev lower bound}) and (\ref{Stdev
upper bound}).

\section{Sample variance penalization versus empirical risk minimization 
\label{Section SVP vs ERM}}

Since empirical Bernstein bounds are observable, have estimation errors
which can be as small as $O\left( 1/n\right) $ for small sample variances,
and can be adjusted to hold uniformly over realistic function classes, they
suggest a method which minimizes the bounds of Corollary \ref{Corollary
empirical Bernstein finite function class} or Theorem \ref{Theorem covering
numbers}. Specifically we consider the algorithm
\begin{equation}
SVP_{\lambda }\left( \mathbf{X}\right) =\arg \min_{f\in \mathcal{F}
}P_{n}\left( f,\mathbf{X}\right) +\lambda \sqrt{\frac{V_{n}\left( f,\mathbf{X
}\right) }{n}},  \label{SVP definition}
\end{equation}
where $\lambda$ is a non-negative parameter. We call this method
sample variance penalization (SVP). Choosing the
regularization parameter $\lambda =0$ reduces the algorithm to
empirical risk minimization (ERM).

It is intuitively clear that SVP will be inferior to ERM if losses
corresponding to better hypotheses have larger variances than the worse
ones. But this seems to be a somewhat unnatural situation. If, on the other
hand, there are some optimal hypotheses of small variance, then SVP should
work well. To make this rigorous we provide a result, which can be used to
bound the excess risk of $SVP_{\lambda }$. Below we use Theorem \ref
{Theorem covering numbers}, but it is clear how the argument is to be
modified to obtain better constants for finite hypothesis spaces.

\begin{theorem}
\label{Theorem excess risk bound}Let $X$ be a random variable with values in
a set $\mathcal{X}$ with distribution $\mu $, and let $\mathcal{F}$ be a
class of hypotheses $f:\mathcal{X\rightarrow }\left[ 0,1\right] $. Fix $%
\delta \in \left( 0,1\right) ,$ $n\geq 2$ and set $\mathcal{M}\left(
n\right) =10\mathcal{N}_{\infty }\left( 1/n,\mathcal{F},2n\right) $ and $%
\lambda =\sqrt{18\ln \left( 3\mathcal{M}\left( n\right) /\delta \right) }$.

Fix $f^{\ast }\in \mathcal{F}$. Then with probability at least $1-\delta $
in the draw of $\mathbf{X}\sim \mu ^{n}$,%
\begin{multline*}
P\left( SVP_{\lambda }\left( \mathbf{X}\right) ,\mu \right) -P\left( f^{\ast
},\mu \right)  \\
\leq \sqrt{\frac{32V\left( f^{\ast },\mu \right) \ln \left( 3\mathcal{M}%
\left( n\right) /\delta \right) }{n}} \\
+\frac{22\ln \left( 3\mathcal{M}\left( n\right) /\delta \right) }{n-1}.
\end{multline*}
\end{theorem}

\begin{proof}
Denote the hypothesis $SVP_{\lambda }\left( \mathbf{X}\right) $ by $\hat{f}$%
. By Theorem \ref{Theorem covering numbers} we have with probability at
least $1-\delta /3$ that%
\begin{eqnarray*}
P\left( \hat{f},\mu \right)  &\leq &P_{n}\left( \hat{f},\mathbf{X}\right)
+\lambda \sqrt{\frac{V_{n}\left( \hat{f},\mathbf{X}\right) }{n}}+\frac{%
15\lambda ^{2}}{18\left( n-1\right) } \\
&\leq &P_{n}\left( f^{\ast },\mathbf{X}\right) +\lambda \sqrt{\frac{%
V_{n}\left( f^{\ast },\mathbf{X}\right) }{n}}+\frac{15\lambda ^{2}}{18\left(
n-1\right) }.
\end{eqnarray*}%
The second inequality follows from the definition of $SVP_{\lambda }$. By
Bennett's inequality (Theorem \ref{Theorem Bernsteins inequality}) we have
with probability at least $1-\delta /3$ that%
\begin{equation*}
P_{n}\left( f^{\ast },\mathbf{X}\right) \leq P\left( f^{\ast },\mu \right) +%
\sqrt{\frac{2V\left( f^{\ast },\mu \right) \ln 3/\delta }{n}}+\frac{\ln
3/\delta }{3n}
\end{equation*}%
and by Theorem \ref{Theorem realvalued} (\ref{Stdev lower bound}) we have
with probability at least $1-\delta /3$ that%
\begin{equation*}
\sqrt{V_{n}\left( f^{\ast },\mathbf{X}\right) }\leq \sqrt{V\left( f^{\ast
},\mu \right) }+\sqrt{\frac{2\ln 3/\delta }{n-1}}.
\end{equation*}%
Combining these three inequalities in a union bound and using $\ln \left( 3%
\mathcal{M}\left( n\right) /\delta \right) \geq 1$ and some other crude but
obvious estimates, we obtain with probability at least $1-\delta $%
\begin{eqnarray*}
P\left( \hat{f},\mu \right)  &\leq &P\left( f^{\ast },\mu \right) +\sqrt{%
\frac{32V\left( f^{\ast },\mu \right) \ln \left( 3\mathcal{M}\left( n\right)
/\delta \right) }{n}} \\
&&+\frac{22\ln \left( 3\mathcal{M}\left( n\right) /\delta \right) }{n-1}.
\end{eqnarray*}
\end{proof}

If we let $f^{\ast }$ be an optimal hypothesis we obtain a bound on the
excess risk. The square-root term in the bound scales with the standard
deviation of this hypothesis, which can be quite small. In particular, if
there is an optimal (minimal risk) hypothesis of zero variance, then the
excess risk of the hypothesis chosen by SVP decays as $\left( \ln \mathcal{M}%
\left( n\right) \right) /n$. In the case of finite hypothesis spaces $%
\mathcal{M}(n)=\left\vert \mathcal{F}\right\vert $ is independent of $n$ and
the excess risk then decays as $1/n$. Observe that apart from the
complexity bound on $\mathcal{F}$ no assumption such as convexity of
the function class or special properties of the loss functions were
needed to derive this result.

To demonstrate a potential competitive edge of SVP over ERM we will now give
a very simple example of this type, where the excess risk of the hypothesis
chosen by ERM is of order $O\left( 1/\sqrt{n}\right) $.

Suppose that $\mathcal{F}$ consists of only two hypotheses $\mathcal{F}%
=\left\{ c_{1/2},b_{1/2+\epsilon }\right\} $. The underlying distribution $\mu $ is
such that $c_{1/2}\left( X\right) =1/2$ almost surely and $b_{1/2+\epsilon
}\left( X\right) $ is a Bernoulli variable with expectation $1/2+\epsilon $,
where $\epsilon \leq 1/\sqrt{8}$. The hypothesis $c_{1/2}$ is optimal and
has zero variance, the hypothesis $b_{1/2+\epsilon }$ has excess risk $%
\epsilon $ and variance $1/4-\epsilon ^{2}$. We are given an i.i.d. sample $%
\mathbf{X}=\left( X_{1},\dots,X_{n}\right) \sim \mu ^{n}$ on which we are to
base the selection of either hypothesis.

It follows from the previous theorem (with $f^{\ast }=c_{1/2}$), that the
excess risk of $SVP_{\lambda }$ decays as $1/n$, for suitably chosen $%
\lambda $. To make our point we need to give a lower bound for the excess
risk of empirical risk minimization. We use the following inequality due to
Slud which we cite in the form given in \cite[p.~363]{Anthony 1999}.

\begin{theorem}
\label{Sluds inequality}Let $B$ be a binomial $\left( n,p\right) $ random
variable with $p\leq 1/2$ and suppose that $np\leq t\leq n\left( 1-p\right)$. 
Then
\begin{equation*}
\Pr \left\{ B>t\right\} \geq \Pr \left\{ Z>\frac{t-np}{\sqrt{np\left(
1-p\right) }}\right\} ,
\end{equation*}
where $Z$ is a standard normal $N\left( 0,1\right) $-distributed random
variable.
\end{theorem}

Now ERM selects the inferior hypothesis $b_{1/2+\epsilon }$ if 
$$
P_{n}\left(b_{1/2+\epsilon },\mathbf{X}\right) <P_{n}\left( c_{1/2},\mathbf{X}\right)
=1/2.
$$ 
We therefore obtain from Theorem \ref{Sluds inequality}, with 
$$
B=n\left( 1-P_{n}\left( b_{1/2+\epsilon }\left( \mathbf{X}\right) \right)
\right),
$$ 
$p=1/2-\epsilon $ and $t=n/2$ that
\begin{eqnarray*}
\Pr \left\{ ERM\left( \mathbf{X}\right) =b_{1/2+\epsilon }\right\}  &=&\Pr
\left\{ P_{n}\left( b_{1/2+\epsilon }\left( \mathbf{X}\right) \right)
<1/2\right\}  \\
&\geq &\Pr \left\{ B>t\right\}  \\
&\geq &\Pr \left\{ Z>\frac{\sqrt{n}\epsilon }{\sqrt{1/4-\epsilon ^{2}}}%
\right\} 
\end{eqnarray*}%
A well known bound for standard normal random variables gives for $\eta >0$%
\begin{eqnarray*}
\Pr \left\{ Z>\eta \right\}  &\geq &\frac{1}{\sqrt{2\pi }}\frac{\eta }{%
1+\eta ^{2}}\exp \left( \frac{-\eta ^{2}}{2}\right)  \\
&\geq &\exp \left( -\eta ^{2}\right) ,\text{ if }\eta \geq 2.
\end{eqnarray*}%
If we assume $n\geq \epsilon ^{-2}$ we have $\sqrt{n}\epsilon /\sqrt{%
1/4-\epsilon ^{2}}\geq 2$, so 
\begin{equation*}
\Pr \left\{ ERM\left( \mathbf{X}\right) =b_{1/2+\epsilon }\right\} \geq \exp
\left( -\frac{n\epsilon ^{2}}{1/4-\epsilon ^{2}}\right) \geq e^{-8n\epsilon
^{2}},
\end{equation*}%
where we used $\epsilon \leq 1/\sqrt{8}$\ in the last inequality. Since this
is just the probability that the excess risk is $\epsilon $ we arrive at the
following statement: For every $n\geq \epsilon ^{-2}$ there exists $\delta $
($=e^{-8n\epsilon ^{2}}$) such that the excess risk of the hypothesis
generated by ERM is at least 
\begin{equation*}
\epsilon =\sqrt{\frac{\ln 1/\delta }{8n}},
\end{equation*}%
with probability at least $\delta $. Therefore the excess risk for ERM
cannot have a faster rate than $O\left( 1/\sqrt{n}\right) $.

This example is of course a very artificial construction, chosen as a simple
illustration. It is clear that the conclusions do not change if we add any
number of deterministic hypotheses with risk larger than $1/2$ (they simply
have no effect), or if we add any number of Bernoulli hypotheses with risk
at least $1/2+\epsilon $ (they just make things worse for ERM).

\begin{figure}[t]
\begin{center}
\includegraphics[width=0.51\textwidth]{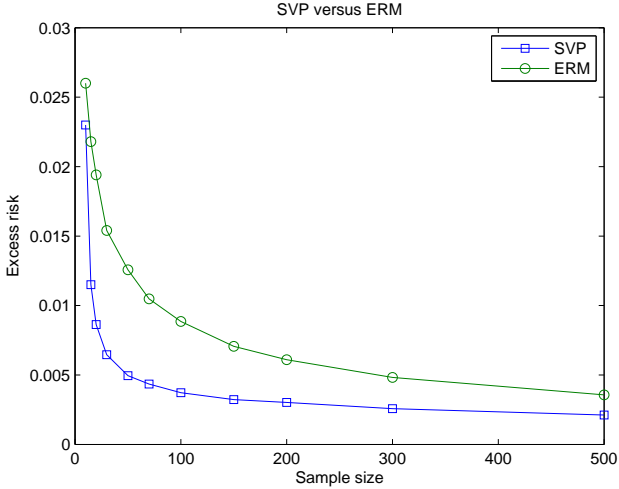}
\end{center}
\caption{Comparison of the excess risks of
the hypotheses returned by ERM (circled line) and SVP with $\protect\lambda =2.5$ (squared line) 
for different sample sizes.}
\label{Experimental Findings} 
\end{figure}

To obtain a more practical
insight into the potential advantages of SVP we have conducted a simple
experiment, where $\mathcal{X}=\left[ 0,1\right] ^{K}$ and the random
variable $X\in \mathcal{X}$ is distributed according to $\prod_{k=1}^{K}\mu
_{a_{k},b_{k}}$ where 
$$\mu _{a,b}=\left( 1/2\right) \left( \delta
_{a-b}+\delta _{a+b}\right). 
$$
Each coordinate $\pi _{k}\left( X\right) $ of 
$X$ is thus a binary random variable, assuming the values $a_{k}-b_{k}$ and $%
a_{k}+b_{k}$ with equal probability, having expectation $a_{k}$ and variance 
$b_{k}^{2}$.

The distribution of $X$ is itself generated at random by selecting the pairs 
$\left( a_{k},b_{k}\right) $ independently: $a_{k}$ is chosen from the
uniform distribution on $\left[ B,1-B\right] $ and the standard deviation $%
b_{k}$ is chosen from the uniform distribution on the interval $\left[ 0,B%
\right] $. Thus $B$ is the only parameter governing the generation of the
distribution.

As hypotheses we just take the $K$ coordinate functions $\pi _{k}$ in $\left[
0,1\right] ^{K}$. Selecting the $k$-th hypothesis then just means that we
select the corresponding distribution $\mu _{a_{k},b_{k}}$. Of course we
want to find a hypothesis of small risk $a_{k}$, but we can only observe $%
a_{k}$ through the corresponding sample, the observation being obscured by
the variance $b_{k}^{2}$.

We chose $B=1/4$ and $K=500$. We tested the algorithm (\ref{SVP
definition}) with $\lambda =0$, corresponding to ERM, and $\lambda
=2.5$. The sample sizes ranged from 10 to 500. We recorded the true
risks of the respective hypotheses generated, and averaged these risks
over 10000 randomly generated distributions. The results are reported
in Figure \ref{Experimental Findings} and show clearly the advantage
of SVP in this particular case. It must however be pointed out that
this advantage, while being consistent, is small compared to the risk
of the optimal hypotheses (around $1/4$).

If we try to extract a practical conclusion from Theorem \ref{Theorem excess
risk bound}, our example and the experiment, then it appears that SVP might
be a good alternative to ERM, whenever the optimal members of the hypothesis
space still have substantial risk (for otherwise ERM would do just as good),
but there are optimal hypotheses of very small variance. These two
conditions seem to be generic for many noisy situations: when the noise
arises from many independent sources, but does not depend too much on any
single source, then the loss of an optimal hypothesis should be sharply
concentrated around its expectation (e.g. by the bounded difference
inequality - see \cite{McDiarmid 1998}), resulting in a small variance. 

\section{Application to sample compression}
\label{sec:SC}

Sample compression schemes \cite{Littlestone 1986} provide an elegant method
to reduce a potentially very complex function class to a finite,
data-dependent subclass. With $\mathcal{F}$ being as usual, assume that some
algorithm $A$ is already specified by a fixed function 
\begin{equation*}
A:\mathbf{X}\in \bigcup_{n=1}^{\infty }\mathcal{X}^{n}\mapsto A_{\mathbf{X}%
}\in \mathcal{F}.
\end{equation*}%
The function $A_{S}$ can be interpreted as the hypothesis chosen by the
algorithm on the basis of the training set $S$, composed with the fixed loss
function. For $x\in \mathcal{X}$ the quantity $A_{S}\left( x\right) $ is
thus the loss incurred by training the algorithm from $S$ and applying the
resulting hypothesis to $x$.

The idea of sample compression schemes \cite{Littlestone 1986} is to
train the algorithm on subsamples of the training data and to use the
remaining data points for testing. A comparison of the different
results then leads to the choice of a subsample and a corresponding
hypothesis. If this hypothesis has small risk, we can say that the
problem-relevant information of the sample is present in the subsample
in a compressed form, hence the name.

Since the method is crucially dependent on the quality of the individual
performance estimates, and empirical Bernstein bounds give tight, variance
sensitive estimates, a combination of sample compression and SVP is
promising. For simplicity we only consider compression sets of a fixed size $%
d$. We introduce the following notation for a subset $I\subset \left\{
1,\dots,n\right\} $ of cardinality $\left\vert I\right\vert =d$.

\begin{itemize}
\item $A_{\mathbf{X}[{I}]}=$ the hypothesis trained with $A$ from the
subsample $\mathbf{X}[{I}]$ consisting of those examples whose indices lie in 
$I$.

\item For $f\in \mathcal{F}$, we let 
$$P_{I^{c}}\left( f\right) =P_{n-d}\left(
f\left( \mathbf{X}[I^{c}]\right) \right)=\frac{1}{n-d}\sum_{i\notin
I}f\left( X_{i}\right),$$ 
the empirical risk of $f$ computed on the
subsample $\mathbf{X}[{I^{c}}]$ consisting of those examples whose
indices do not lie in $I$.

\item For $f\in \mathcal{F}$, we let 
\begin{eqnarray}
\nonumber
V_{I^{c}}(f)  & \hspace{-.21truecm}= \hspace{-.21truecm}&V_{n-d}\left(
f\left( \mathbf{X}[{I^{c}}]\right) \right)~~~ \\ 
\nonumber
~ & 
\hspace{-.21truecm} = \hspace{-.21truecm}&\frac{1}{2(n-d)(n-d-1)}
\hspace{-.1truecm} \sum_{i,j\notin I}\left( f\left( X_{i}\right) -f\left( X_{j}\right)
\right) ^{2}\hspace{-.15truecm},~~~
\end{eqnarray}
the sample variance of $f$ computed on $\mathbf{X}[{I^{c}}]$.

\item $\mathcal{C}=$ the collection of subsets $I\subset \left\{
1,\dots,n\right\} $ of cardinality $\left\vert I\right\vert =d$. 
\end{itemize}

\noindent With this notation we define our sample compression scheme as%
\begin{eqnarray*}
SVP_{\lambda }\left( \mathbf{X}\right)  &=&A_{\mathbf{X}[{\hat{I}}]} \\
\hat{I} &=&\arg \min_{I\in \mathcal{C}}P_{I^{c}}\left( A_{\mathbf{X}[
{I}]}\right) +\lambda \sqrt{V_{I^{c}}\left( A_{\mathbf{X}[{I}]}\right) }.
\end{eqnarray*}%
As usual, $\lambda =0$ gives the classical sample compression schemes. The
performance of this algorithm can be guaranteed by the following result.

\begin{theorem}
With the notation introduced above fix $\delta \in \left( 0,1\right) ,$ $%
n\geq 2$ and set $\lambda =\sqrt{2\ln \left( 6\left\vert \mathcal{C}%
\right\vert /\delta \right) }$. Then with probability at least $1-\delta $
in the draw of $\mathbf{X}\sim \mu ^{n}$, we have for every $I^{\ast }\in 
\mathcal{C}$%
\begin{multline*}
P\left( SVP_{\lambda }\left( \mathbf{X}\right) ,\mu \right) -P\left( A_{%
\mathbf{X}[{I^*}]},\mu \right)  \\
\leq \sqrt{\frac{8V\left( A_{\mathbf{X}[{I^{\ast }}]},\mu \right) \ln \left(
6\left\vert \mathcal{C}\right\vert /\delta \right) }{n-d}}+\frac{14\ln
\left( 6\left\vert \mathcal{C}\right\vert /\delta \right) }{3\left(
n-d-1\right) }
\end{multline*}
\end{theorem}

\begin{proof} Use a union bound and Theorem \ref{Theorem empirical Bernstein bound degree
1} to obtain an empirical Bernstein bound uniformly valid over all $A_{%
\mathbf{X}[{I}]}$ with $I\in \mathcal{C}$ and therefore also valid for $%
SVP_{\lambda }\left( \mathbf{X}\right) $. Then follow the proof of Theorem %
\ref{Theorem excess risk bound}. Since now $I^{\ast }\in \mathcal{C}$ is
chosen \textit{after} seeing the sample, uniform versions of Bennett's
inequality and Theorem \ref{Theorem realvalued} (\ref{Stdev lower bound})
have to be used, and are again readily obtained with union bounds over $%
\mathcal{C}$.
\end{proof}

The interpretation of this result as an excess risk bound is more subtle
than for Theorem \ref{Theorem excess risk bound}, because the optimal
hypothesis is now sample-dependent. If we define%
\begin{equation*}
I^{\ast }=\arg \min_{I\in \mathcal{C}}P\left( A_{\mathbf{X}[{I}]},\mu \right)
,
\end{equation*}%
then the theorem tells us how close we are to the choice of the optimal
subsample. This will be considerably better than what we get from Hoeffding's inequality if the variance $V\left( A_{\mathbf{X}[{I^{\ast }}]},\mu
\right) $ is small and sparse solutions are sought in the sense that $d/n$
is small (observe that $\ln \left\vert \mathcal{C}\right\vert \leq d\ln
\left( ne/d\right) $).

This type of relative excess risk bound is of course more useful if the
minimum $P\left( A_{\mathbf{X}[{I^{\ast }}]},\mu \right) $ is close to some
true optimum arising from some underlying generative model. In this case we
can expect the loss $A_{\mathbf{X}[{I^{\ast }}]}$ to behave like a noise
variable centered at the risk $P\left( A_{\mathbf{X}[{I^{\ast }}]},\mu
\right) $. If the noise arises from many independent sources, each of which
makes only a small contribution, then $A_{\mathbf{X}[{I^{\ast }}]}$ will be
sharply concentrated and have a small variance $V\left( A_{\mathbf{X}
[{I^{\ast }}]},\mu \right) $, resulting in tight control of the excess
risk.\bigskip

\section{Conclusion}
We presented sample variance penalization as a potential alternative to
empirical risk minimization and analyzed some of its statistical properties
in terms of empirical Bernstein bounds and concentration properties of the
empirical standard deviation. The promise of our method is that, 
in simple but perhaps practical scenarios the excess risk of our method 
is guaranteed to be substantially better than that of empirical 
risk minimization. 

The present work raises some questions. Perhaps the most pressing
issue is to find an efficient implementation of the method, to deal
with the fact that sample variance penalization is non-convex in many
situations when empirical risk minimization is convex, and to compare
the two methods on some real-life data sets. Another important issue
is to further investigate the application of empirical Bernstein
bounds to sample compression schemes.

\end{document}